# Learning ELM network weights using linear discriminant analysis

Philip de Chazal[1], Jonathan Tapson[1] and André van Schaik[1]

[1]The MARCS Institute, University of Western Sydney, Penrith NSW 2751, Australia

`{p.dechazal, j.tapson, a.vanschaik}@uws.edu.au`

**Abstract.** We present an alternative to the pseudo-inverse method for determining the hidden to output weight values for Extreme Learning Machines performing classification tasks. The method is based on linear discriminant analysis and provides Bayes optimal single point estimates for the weight values.

**Keywords:** Extreme learning machine, Linear discriminant analysis, Hidden to output weight optimization, MNIST database

## 1   Introduction

The Extreme Learning Machine (ELM) is a multi-layer feedforward neural network that offers fast training and flexible non-linearity for function and classification tasks. Its principal benefit is that the network parameters are calculated in a single pass during the training process [1]. In its standard form it has an input layer that is fully connected to a hidden layer with non-linear activation functions. The hidden layer is fully connected to an output layer with linear activation functions. The number of hidden units is often much greater than the input layer with a fan-out of 5 to 20 hidden units per input frequently used. A key feature of ELMs is that the weights connecting the input layer to the hidden layer are set to random values. This simplifies the requirements for training to one of determining the hidden to output unit weights, which can be achieved in a single pass. By randomly projecting the inputs to a much higher dimensionality, it is possible to find a hyperplane which approximates a desired regression function, or represents a linear separable classification problem [2].

A common way of calculating the hidden to output weights is to use the Moore-Penrose pseudo-inverse applied to the hidden layer outputs using labelled training data. In this paper we present an alternative method for hidden to output weight calculation for networks performing classification tasks. The advantage of our method over the pseudo-inverse method is that the weights are the best single point estimates from a Bayesian perspective for a linear output stage. Using the same network architecture and same random values for the input to hidden layer weights, we applied the two weight calculation methods to the MNIST database and demonstrated that our method offers a performance advantage.

## 2  Methods

If we consider a particular sample of input data $\mathbf{x}_k \in \mathbb{R}^{L \times 1}$ where $k$ is a series index and $K$ is the length of the series, then the forward propagation of the local signals through the network can be described by:

$$y_{n,k} = \sum_{m=1}^{M} w_{nm}^{(2)} g\left( \sum_{l=1}^{L} w_{ml}^{(1)} x_{l,k} \right) \qquad (1)$$

Where $\mathbf{y}_k \in \mathbb{R}^{N \times 1}$ is the output vector corresponding to the input vector $\mathbf{x}_k$, $l$ and $L$ are the input layer index and number of input features respectively, $m$ and $M$ are the hidden layer index and number of hidden units respectively, and $n$ and $N$ are the output layer index and number of output units respectively. $w^{(1)}$ and $w^{(2)}$ are the weights associated with the input to hidden layer and the hidden to output layer linear sums respectively. $g(\ )$ is the hidden layer non-linear activation function.

With ELM, $w^{(1)}$ are assigned randomly which simplifies the training requirements to task of optimisation of the $w^{(2)}$ only. The choice of linear output neurons further simplifies the optimisation problem of $w^{(2)}$ to a single pass algorithm.

The weight optimisation problem for $w^{(2)}$ can be stated as

$$y_{n,k} = \sum_{m=1}^{M} w_{nm}^{(2)} a_{m,k} \text{ where } a_{m,k} = g\left( \sum_{l=1}^{L} w_{ml}^{(1)} x_{l,k} \right). \qquad (2)$$

We can restate this as matrix equation by using $\mathbf{W} \in \mathbb{R}^{N \times M}$ with elements $w_{nm}^{(2)}$, and $\mathbf{A} \in \mathbb{R}^{M \times K}$ in which each column contains outputs of the hidden unit at one instant in the series $\mathbf{a}_k \in \mathbb{R}^{M \times 1}$, and the output $\mathbf{Y} \in \mathbb{R}^{N \times K}$ where each column contains output of the network at one instance in the series as follows:

$$\mathbf{Y} = \mathbf{W}\mathbf{A}. \qquad (3)$$

The optimisation problem involves determining the matrix $\mathbf{W}$ given a series of desired outputs for $\mathbf{Y}$ and a series of hidden layer outputs $\mathbf{A}$.
We represent the desired outputs for $\mathbf{y}_k$ using the target vectors $\mathbf{t}_k \in \mathbb{R}^{N \times 1}$ where $t_{n,k}$ has value 1 in the row corresponding to the desired class and 0 for the other N-1 elements. For example $\mathbf{t}_k = [0,1,0,0]^T$ indicates the desired target is class 2 (of four classes). As above we can restate the desired targets using a matrix $\mathbf{T} \in \mathbb{R}^{N \times K}$ where each column contains the desired targets of the network at one instance in the series. Substituting $\mathbf{T}$ in for the desired outputs for $\mathbf{Y}$, the optimization problem involves solving the following linear equation for $\mathbf{W}$:

$$\mathbf{T} = \mathbf{WA}. \tag{4}$$

### 2.1 Output weight calculation using the pseudo-inverse

In ELM literature $\mathbf{W}$ is often determined by the taking the Moore-Penrose pseudo-inverse $\mathbf{A}^+ \in \mathbb{R}^{K \times M}$ of $\mathbf{A}$ [3]. If the rows of are $\mathbf{A}$ are linearly independent (which normally true if K>M) then $\mathbf{W}$ maybe calculated using

$$\mathbf{W} = \mathbf{TA}^+ \text{ where } \mathbf{A}^+ = \mathbf{A}^T \left( \mathbf{AA}^T \right)^{-1}. \tag{5}$$

This minimises the sum of square error between networks outputs $\mathbf{Y}$ and the desired outputs $\mathbf{T}$, i.e.

$$\mathbf{A}^+ \text{ minimises } \|\mathbf{Y} - \mathbf{T}\|_2 = \sum_{k=1}^{K} \sum_{n=1}^{N} \left( y_{n,k} - t_{n,k} \right)^2 \tag{6}$$

We refer to the pseudo-inverse method for output weight calculation as PI-ELM. We note that in cases where the classification problem is ill-posed it may be necessary to regularize this solution, using standard methods such as Tikhonov regularization (ridge regression).

### 2.2 Output weight calculation using linear discriminant analysis

In this paper we develop an alternative approach to estimating $\mathbf{W}$ based on a maximum likelihood estimator assuming a linear model. We refer to it as the LDA-ELM method as it is equivalent to applying linear discriminant analysis to the hidden layer outputs. Our presentation is based on the notation of Ripley [4].

For an *N*-class problem Bayes' rule states that the posterior probability of the *n*th class $p_n$ is related to its prior probability $\pi_k$ and its class density function $f_n(\mathbf{d}, \boldsymbol{\theta}_n)$ by

$$p_n = \frac{\pi_n f_n(\mathbf{d}, \boldsymbol{\theta}_n)}{\sum_{z=1}^{N} \pi_z f_z(\mathbf{d}, \boldsymbol{\theta}_z)} \tag{7}$$

where $\mathbf{d}$ is the input data vector (in our case the hidden layer output), and $\boldsymbol{\theta}_n$ are the parameters of the class density function.

The class densities are modelled with a multi-variate Gaussian model with common covariance $\Sigma$ and class dependent mean vectors $\mu_k$. Given an input vector $\mathbf{a}_k$ the class density is

$$f_n(\mathbf{a}_k, \theta_n = \mu_n, \Sigma) = (2\pi)^{-\frac{M}{2}} |\Sigma|^{-\frac{1}{2}} \exp\left[-\tfrac{1}{2}(\mathbf{a}_k - \mu_n)^T \Sigma^{-1}(\mathbf{a}_k - \mu_n)\right] \tag{8}$$

We set the dimension of the Gaussian model equal to the number of hidden units so that $\mathbf{a}_k$ is as defined above for the hidden unit output and hence $\Sigma \in \mathbb{R}^{M \times M}$ and $\mu_n \in \mathbb{R}^{M \times 1}$.

To begin with, the training data is partitioned according to the class membership so that we have $K = \sum_{n=1}^{N} K_n$ labelled data vectors of hidden unit outputs, $\mathbf{a}_k^{(n)} \in \mathbb{R}^{M \times 1}, k = 1..K_n$ where all members $\mathbf{a}^{(n)}$ belong to class $n$.

For a given set of hidden unit output data and class membership a likelihood function $l(\theta)$ is formed using

$$l(\theta) = l(\theta_1, \theta_2, ..., \theta_N) = \prod_{n=1}^{N} \prod_{k=1}^{K_n} \pi_n f_n(\mathbf{a}_k^{(n)}, \theta_n) \tag{9}$$

Our aim is to find values of $\theta_n$ that maximise $l(\theta)$ for given set of training data. Equivalently we can maximise the value of the log-likelihood:

$$L(\theta_1, \theta_2, ..., \theta_N) = \log(l(\theta_1, \theta_2, ..., \theta_N)) = \sum_{n=1}^{N} \sum_{k=1}^{K_n} \log(f_n(\mathbf{a}_k^{(n)}, \theta_n)) + \sum_{n=1}^{N} K_n \log(\pi_n) \tag{10}$$

Substituting our multi-variate Gaussian model for $f_n(\mathbf{a}_k, \theta_n)$ we get

$$L(\theta_1, \theta_2, ..., \theta_N) = L(\mu_1, ..., \mu_N, \Sigma) = \\ \sum_{n=1}^{N} \sum_{k=1}^{K_n} \left(-\tfrac{M}{2} \log(2\pi) - \tfrac{1}{2} \log(|\Sigma|) - \tfrac{1}{2}(\mathbf{a}_k^{(n)} - \mu_n)^T \Sigma^{-1}(\mathbf{a}_k^{(n)} - \mu_n)\right) + \sum_{n=1}^{N} K_n \log(\pi_n). \tag{11}$$

This is maximized when

$$\mu_n = \sum_{k=1}^{K_n} \mathbf{a}_k^{(n)} \Big/ K_n, \text{ and } \Sigma = \sum_{n=1}^{N} \sum_{k=1}^{K_n} (\mathbf{a}_k^{(n)} - \mu_k)(\mathbf{a}_k^{(n)} - \mu_k)^T \Big/ K. \tag{12}$$

Having determined the $\mu_n$'s and $\Sigma$ from the training data we now need to find the values for $\mathbf{W}$. We begin by substituting (8) into (7), bringing the $\pi_n$ into exponential function and removing the common numerator and denominator term $(2\pi)^{-\frac{M}{2}}|\Sigma|^{-\frac{1}{2}}$, giving us

$$p_n = \frac{\exp\left[-\frac{1}{2}(\mathbf{a}_k - \boldsymbol{\mu}_n)^T \Sigma^{-1}(\mathbf{a}_k - \boldsymbol{\mu}_n) + \log(\pi_n)\right]}{\sum_{z=1}^{N} \exp\left[-\frac{1}{2}(\mathbf{a}_k - \boldsymbol{\mu}_z)^T \Sigma^{-1}(\mathbf{a}_k - \boldsymbol{\mu}_z) + \log(\pi_z)\right]}. \tag{13}$$

After expanding the $-\frac{1}{2}(\mathbf{a}_k - \boldsymbol{\mu}_n)^T \Sigma^{-1}(\mathbf{a}_k - \boldsymbol{\mu}_n)$ terms and removing the $-\frac{1}{2}\mathbf{a}_k^T \Sigma^{-1}\mathbf{a}_k$ from the numerator and denominator we get

$$p_n = \frac{\exp(y_n)}{\sum_{a=1}^{N} \exp(y_a)} \tag{14}$$

where

$$y_n = \log(\pi_n) + \boldsymbol{\mu}_n^T \Sigma^{-1}\mathbf{a}_k - \frac{1}{2}\boldsymbol{\mu}_n^T \Sigma^{-1}\boldsymbol{\mu}_n \tag{15}$$

Classification is performed by choosing the class with the highest value of $p_n$. As $p_n$ in (14) is a monotonic function of $y_n$ in (15) we can use either function when deciding our final class. We choose to use $y_n$ defined in (15) as it is a linear function of the input data vector $\mathbf{a}_k$ and it can be used to determine $\mathbf{W}$ for our network as follows:

$$\mathbf{W} = \begin{bmatrix} \log(\pi_1) - \frac{1}{2}\boldsymbol{\mu}_1^T \Sigma^{-1}\boldsymbol{\mu}_1 & \log(\pi_2) - \frac{1}{2}\boldsymbol{\mu}_2^T \Sigma^{-1}\boldsymbol{\mu}_2 & \cdots & \log(\pi_N) - \frac{1}{2}\boldsymbol{\mu}_N^T \Sigma^{-1}\boldsymbol{\mu}_N \\ \boldsymbol{\mu}_1^T \Sigma^{-1} & \boldsymbol{\mu}_2^T \Sigma^{-1} & \cdots & \boldsymbol{\mu}_N^T \Sigma^{-1} \end{bmatrix}. \tag{16}$$

Note that $\mathbf{W} \in \mathbb{R}^{N \times M+1}$, as a constant term has been introduced into the hidden to output layer weights (the first row of $\mathbf{W}$). If we want to determine the posterior probabilities then we use (14) applied to the network outputs.

**Summary of method**

In summary calculating $\mathbf{W}$ proceeds as follows

(i) Partitioned the hidden unit output data according to the class membership so that we have $K = \sum_{n=1}^{N} K_n$ labelled data vectors, $\mathbf{a}_k^{(n)} \in \mathbb{R}^{L \times 1}, k = 1..K_n$ where all members $\mathbf{a}^{(n)}$ belong to class $n$

(ii) Calculate $\boldsymbol{\mu}_n = \sum_{k=1}^{K_n} \mathbf{a}_k^{(n)} / K_n$ and $\Sigma = \sum_{n=1}^{N} \sum_{k=1}^{K_n} (\mathbf{a}_k^{(n)} - \boldsymbol{\mu}_k)(\mathbf{a}_k^{(n)} - \boldsymbol{\mu}_k)^T / K$

(iii) Set the prior probabilities $\pi_n$

(iv) Calculate

$$\mathbf{W} = \begin{bmatrix} \log(\pi_1) - \frac{1}{2}\boldsymbol{\mu}_1^T \boldsymbol{\Sigma}^{-1} \boldsymbol{\mu}_1 & \log(\pi_2) - \frac{1}{2}\boldsymbol{\mu}_2^T \boldsymbol{\Sigma}^{-1} \boldsymbol{\mu}_2 & \cdots & \log(\pi_N) - \frac{1}{2}\boldsymbol{\mu}_N^T \boldsymbol{\Sigma}^{-1} \boldsymbol{\mu}_N \\ \boldsymbol{\mu}_1^T \boldsymbol{\Sigma}^{-1} & \boldsymbol{\mu}_2^T \boldsymbol{\Sigma}^{-1} & \cdots & \boldsymbol{\mu}_N^T \boldsymbol{\Sigma}^{-1} \end{bmatrix}$$

To classify new data we
  (i) Calculate the network output **y** in response to the hidden layer output **a** is

$$\mathbf{y} = \mathbf{W} \begin{bmatrix} 1 \\ \mathbf{a} \end{bmatrix}$$

  (ii) (Optional) Calculate the posterior probabilities

$$p_n = \frac{\exp(y_n)}{\sum_{a=1}^{N} \exp(y_a)}$$

  (iii) The final decision of the network is the output with the highest value of $y_n$ or, equivalently, $p_n$.

**Combining Classifiers**

Equation (14) provides an easy way to combine the outputs of multiple classifiers. Once the posterior probabilities are calculated for each class for each classifier we can form a combined posterior probability and choose the class with the highest combined posterior probability. There are any schemes for doing this [5] with unweighted averaging across the posterior probability outputs being one of the most simple schemes.

## 3   Experiments

We applied the LDA-ELM and PI-ELM weight calculation method to the MNIST handwritten digit recognition problem [6]. Authors JT and AvS have previously reported good classification results using ELM on this database [2]. The database has 60,000 training and 10,000 testing examples. Each example is a 28*28 pixel 256 level grayscale image of a handwritten digit between 0 and 9. The 10 classes are approximately equally distributed in the training and testing sets.

The ELM algorithms were applied directly to the unprocessed images and we trained the networks by providing all data in batch mode. The random values for the input layer weights were uniformly distributed between -0.5 and 0.5. The prior probabilities for the 10 classes for LDA-ELM were each set to 0.1.

In order to perform a direct comparison of the two methods we used the following protocol:

For fan-out of 1 to 20 hidden units per input, repeat 200 times
  (i)  Assign random values to the input layer weights and determine the hidden layer outputs for the 60,000 training data examples.
  (ii) Determine PI-ELM network weights using data from (i).
  (iii) Determine LDA-ELM network weights using data from (i).
  (iv) Evaluate both networks on the 10,000 test data examples and store results.

We averaged the results for the 200 repeats of the experiment for each fan-out and compared the misclassification rates. These results are shown in Fig. 1 and Table 1.

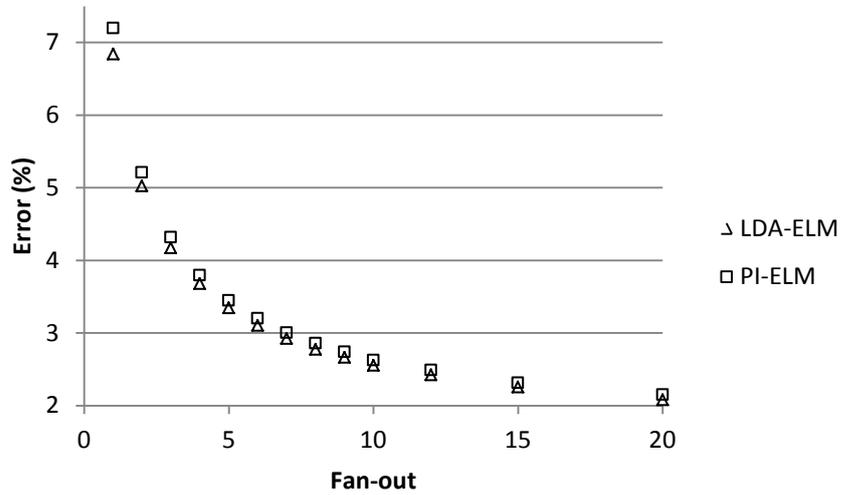

**Fig. 1.** The error rate of the LDA-ELM and the PI-ELM on the MNIST database for fan-out varying between 1 and 20. All results at each fan-out are averaged from 200 repeats of the experiment.

The results show that the LDA-ELM outperforms the PI-ELM at every fan-out value. The average performance benefit was a 3.1% decrease in the error rate of LDA-ELM with a larger benefit at smaller fan-out values. Table 2 below shows that there is little extra computational requirement for the LDA-ELM method.

**Table 1.** The error rate (%) and percentage improvement of the LDA-ELM over the PI-ELM on the MNIST database. Results averaged from 200 repeats.

| | Fanout | | | | | | | | | | | | |
|---|---|---|---|---|---|---|---|---|---|---|---|---|---|
| | 1 | 2 | 3 | 4 | 5 | 6 | 7 | 8 | 9 | 10 | 12 | 15 | 20 |
| PI-ELM | 7.20 | 5.21 | 4.32 | 3.80 | 3.45 | 3.20 | 3.00 | 2.86 | 2.74 | 2.63 | 2.49 | 2.31 | 2.15 |
| LDA-ELM | 6.84 | 5.03 | 4.17 | 3.68 | 3.35 | 3.11 | 2.92 | 2.78 | 2.66 | 2.55 | 2.42 | 2.25 | 2.08 |
| % improvement | 4.9 | 3.5 | 3.3 | 3.1 | 2.9 | 3.0 | 2.6 | 2.9 | 2.8 | 2.7 | 2.6 | 2.6 | 3.3 |

**Table 2.** Computation times (in seconds). The elapsed time is shown for training the PI-ELM and LDA-ELM networks on the 60,000 images from MNIST database and testing on the 10,000 images using MATLAB R2012a code running on 2012 Sony Vaio Z series laptop with an Intel i7-2640M 2.8GHz processor and 8GB RAM.

| | Fan-out | | | | | | | | | | | | |
|---|---|---|---|---|---|---|---|---|---|---|---|---|---|
| | 1 | 2 | 3 | 4 | 5 | 6 | 7 | 8 | 9 | 10 | 12 | 15 | 20 |
| PI-ELM | 6.2 | 13.9 | 24.9 | 37.8 | 53.3 | 68.5 | 88.2 | 111 | 136 | 162 | 228 | 339 | 630 |
| LDA-ELM | 6.2 | 13.9 | 25.2 | 38.1 | 54.0 | 69.5 | 90.6 | 113 | 140 | 167 | 238 | 357 | 702 |

The last experiment we performed investigated combining multiple networks using the LDA-ELM by averaging posterior probabilities. We investigated using an ensemble number between 1 and 20 and repeated the training and testing 10 times at each ensemble number. We then averaged the results which are shown below in Fig. 2.

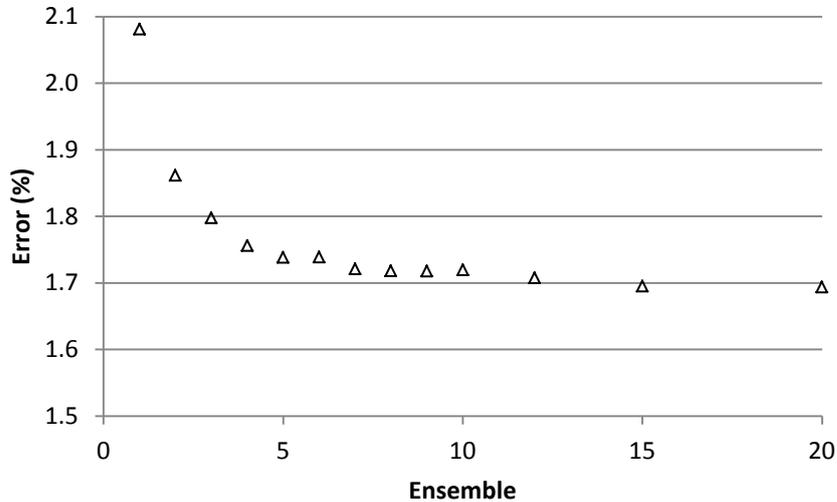

**Fig. 2.** The error rate of the LDA-ELM on the MNIST database at a fan-out of 20 with the ensemble number varying between 1 and 20. The result at each ensemble number is averaged from 10 repeats of the experiment.

The results shown in Fig. 2 demonstrate the benefit of combining multiple LDA-ELM networks on the MNIST database. Combining two networks reduced the error rate from 2.08% to 1.86% and adding more networks further reduced the error. The best error rate was 1.69% achieved when 20 networks were combined.

## 4   Discussion

The results on the MNIST database shown in Fig. 1 suggest that there is a performance benefit to be gained by using the LDA-ELM output weight calculation over the PI-ELM method. As there is only a small extra computation overhead we believe it is a viable alternative to the pseudo-inverse method especially at small fan-out values.

Another benefit of the LDA-ELM is the ability to combine outputs from networks by combining the posterior probabilities estimates of the individual networks. When we applied this to the MNIST database we were able to reduce the error rate to 1.7%. This result is comparable to the best performance of most other 2 and 3 layer neural networks processing the raw data [7]. Further work will include comparing the two weight calculation methods on other publicly available databases such as abalone and iris data sets [8].

## 5     Conclusion

We have presented a new method for weight calculation for hidden to output weights for ELM networks performing classification tasks. The method is based on linear discriminant analysis and requires a modest amount of extra calculation time compared to the pseudo-inverse method (<12% for a fan-out ≤ 20). When applied to the MNIST database the average misclassification rate improvement was 3.1% in comparison to the pseudo-inverse method for identically configured and initialized networks.